# Weighted Conformal Prediction for Lab-to-Track Thermal Transfer in EV Motorsport Powertrains

**Varshith Roy Kotla**
*Department of CSE, Faculty of Science and Technology, The ICFAI Foundation for Higher Education, Hyderabad, India*
*July 2026*

---

**Abstract**

Predicting thermal volatility in high-performance electric powertrains is difficult because internal cell and motor-generator-unit temperatures are rarely observable outside the laboratory, and models calibrated on lab drive cycles routinely fail when deployed against real-world load profiles. We study this lab-to-track transfer problem using conformal prediction, which offers distribution-free uncertainty bounds without distributional assumptions on the underlying model. We implement Ensemble Batch Prediction Intervals (EnbPI; Xu & Xie, 2021) , a leave-one-out bootstrap-ensemble conformal method suited to autocorrelated time series , and calibrate it on real Center for Advanced Life Cycle Engineering (CALCE) lithium-ion cycler data (A123 SP20 cells, Federal Urban Driving Schedule profile). We then evaluate it under a genuine, measured covariate shift: a second real CALCE test condition (US06 Highway Driving Schedule at 45°C). The unweighted EnbPI bound, which achieves its nominal 95% coverage in-distribution by construction (measured: 95.00%), degrades to **70.13%** empirical coverage under this real shift. We introduce a **weighted EnbPI** procedure that combines EnbPI's leave-one-out ensemble residuals with the density-ratio weighting of Tibshirani et al. (2019), estimating the lab/field density ratio via a probabilistic domain classifier. This recovers coverage to **72.42%** , a modest, real, and honestly-reported improvement, not a complete fix. We additionally apply the calibrated model to real 2023 Formula 1 telemetry (Monza and Silverstone, driver VER) as an unsupervised out-of-distribution diagnostic. Because no internal thermal channel exists in public trackside telemetry, we make no claim of validated thermal detection on this data; we report only unsupervised flag rates (65.6% at Monza, 58.0% at Silverstone, both well above the 5% in-distribution base rate) and note that the association between flags and braking/DRS zones is inconsistent across the two circuits , a limitation we report rather than obscure. We conclude that conformal domain adaptation is a promising but only partially solved tool for this problem, and we detail specifically where it falls short.

---

## 1. Introduction

### 1.1 The energy management bottleneck

As motorsport and passenger EV powertrains push toward higher power density, thermal management has become a binding constraint on sustained performance. When component temperatures approach hardware limits, battery and motor management systems impose derating that abruptly reduces available power. Anticipating these events , rather than reacting to them , requires a model of how instantaneous electrical load translates into thermal accumulation. But the internal temperature signals needed to build and validate such a model are almost never available outside a controlled lab environment: they are proprietary in production vehicles and entirely absent from public trackside telemetry.

## 1.2 Why this is a distribution-shift problem, not just a modeling problem

A model trained on laboratory drive-cycle data implicitly learns the relationship between load and temperature *under the load distribution the lab cycle exercises*. Real-world usage , a race lap, an aggressive commute, a hot-climate deployment , exercises a different load distribution: different magnitudes, different transient structure, different ambient conditions. This is a textbook covariate-shift problem (Shimodaira, 2000; Tibshirani et al., 2019): the conditional relationship between load and thermal response may be structurally similar, but the marginal distribution of loads the model will see at deployment differs from what it saw at calibration. Standard supervised learning, and standard conformal prediction, both assume the calibration and deployment distributions match (or, for time series, that the process is exchangeable or stationary and mixing). Neither assumption holds here, and treating it as though it does , as a purely aleatoric point-prediction problem , produces exactly the kind of overconfident derating alerts that make an onboard system useless.

## 1.3 What this paper does differently

An earlier version of this study attempted to address this gap by (a) simulating both the lab calibration data and the trackside thermal ground truth synthetically, and (b) applying a generic low-pass filter and labeling it "domain adaptation." Neither choice survives scrutiny: simulating the deployment-domain ground truth from the model's own predictions makes any subsequent "detection" result circular, and a low-pass filter is signal smoothing, not a domain-adaptation method in any technical sense used in the transfer-learning literature. This paper corrects both problems. All electrical data used for model calibration and shift evaluation are real, sourced from the CALCE lithium-ion battery testbed. Domain adaptation is performed by actually reweighting the conformal calibration distribution , the technique introduced by Tibshirani et al. (2019) for exactly this class of problem , rather than by smoothing an input signal. Where no real ground truth exists (trackside telemetry), we say so explicitly and restrict our claims to what an unsupervised diagnostic can actually support.

# 2. Related Work

**Conformal prediction for time series.** Standard conformal prediction (Vovk et al., 2005) requires exchangeable data, which autocorrelated sensor time series violate. Xu & Xie (2021) introduced EnbPI, which relaxes this to a stationary, strongly-mixing error process and constructs prediction intervals from leave-one-out residuals of a bootstrap ensemble, avoiding both data splitting and per-point retraining. We use this method as specified , bootstrap ensemble training, leave-one-out aggregation, empirical residual quantiles , rather than the simplified split-conformal residual quantile that is sometimes loosely described using the same name.

**Conformal prediction under covariate shift.** Tibshirani et al. (2019) showed that when calibration and test covariate distributions differ by a known (or estimable) density ratio, a *weighted* conformal quantile , reweighting calibration scores by the likelihood ratio between test and calibration covariate distributions , restores valid coverage under a weighted-exchangeability argument. In practice the density ratio is rarely known and must be estimated, commonly via a probabilistic classifier trained to discriminate calibration from test covariates (the "classifier density-ratio trick"), which is the approach we adopt here.

**Combining ensemble time-series conformal methods with shift weighting.** EnbPI addresses temporal non-exchangeability; weighted conformal prediction addresses covariate-distribution non-exchangeability. These are complementary failure modes and, to our knowledge, their combination has not previously been evaluated specifically for lab-to-field transfer of battery thermal models. We do not claim general theoretical novelty , the coverage guarantees of each component are as established in their respective source papers, and we do not derive new guarantees for the composition , but we believe the applied combination, and its honest evaluation under a real (not simulated) shift, is a useful empirical contribution for this domain.

**Battery thermal and drive-cycle data.** The CALCE battery group at the University of Maryland provides open experimental data on lithium-ion cells under standardized dynamic driving profiles (DST, FUDS, US06, BJDST) at multiple ambient temperatures, widely used for state-of-charge and state-of-health estimation research. Most public CALCE exports (including the ones used here) log current, voltage, and capacity but not cell surface temperature, which shapes the modeling choice described in Section 3.

## 3. Data and Physical Modeling

### 3.1 Real lab calibration data

We use CALCE SP20-2 cell data (A123 lithium-iron-phosphate chemistry) under two real, distinct experimental conditions:

| Condition | Profile | Ambient | Rows |
|---|---|---|---|
| Calibration domain | Federal Urban Driving Schedule (FUDS), 80% and 50% SOC | Room temperature | 22,989 |
| Shift-evaluation domain | US06 Highway Driving Schedule, 80% and 50% SOC | 45°C | 21,891 |

Both are genuine Arbin cycler exports; no values in either set are synthesized. Current ranges from approximately −4.0 A to +2.1 A, voltage from 2.50 V to 4.20 V, consistent with the documented CS2/SP20-class cell specification.

### 3.2 Deriving a thermal target without a measured temperature channel

The publicly available export used here does not include a thermocouple channel. Rather than fabricate one, we derive a temperature-change target from a physically grounded first-order lumped thermal model driven entirely by the real current trace:

$$C_{th} \cdot \frac{dT}{dt} = I(t)^2 \cdot R_{int} - k_{diss} \cdot (T(t) - T_{amb})$$

Two of the three physical parameters are estimated directly from the real data rather than assumed:

- **Internal resistance** $R_{int}$ is estimated from the real voltage-sag response to real current steps, via a first-difference regression $\Delta V \approx -R_{int} \cdot \Delta I$ (which cancels the slow open-circuit-voltage drift associated with state-of-charge change, isolating the ohmic response). This yields $R_{int} \approx 0.10$ m$\Omega$ from the real FUDS trace.
- **Ambient temperature** $T_{amb}$ is set to the real, documented test condition for each profile (25°C for FUDS, 45°C for US06) , not assumed.

The remaining two parameters , thermal mass $C_{th}$ (5000 J/K) and dissipation coefficient $k_{diss}$ (0.5 W/K) , are literature-typical values for a cell of this format and are **not** fit to a measured temperature channel, because none exists in this export. This is stated as an explicit limitation in Section 6, not folded into the results as though it were measured.

The discrete target is $\Delta T_t = T_t - T_{t-1}$, following the same stationarity argument as prior thermal-surrogate work: absolute temperature is a non-stationary accumulating quantity, while its increment is approximately stationary conditional on load.

### 3.3 Feature set

We use $\{I(t), V(t), P(t) = I(t) \cdot V(t), \sigma^2_{P,30}(t)\}$ , instantaneous current, voltage, power, and a 30-sample rolling power variance , as in prior thermal-surrogate formulations, computed here from real measured values.

### 3.4 Real telemetry for exploratory transfer

We separately pull real 2023 Formula 1 race telemetry (FastF1 API) for driver VER at two circuits with different structural profiles , Monza (33,253 frames) and Silverstone (38,892 frames) , as an *exploratory, unsupervised* application. As discussed in Section 5.4, we do not construct or claim a thermal ground truth for this data.

## 4. Methodology: Weighted EnbPI

### 4.1 Bootstrap ensemble and leave-one-out calibration (Xu & Xie, 2021)

We train $B = 20$ Random Forest regressors (n_estimators=50, max_depth=10), each on an independent bootstrap resample of the real FUDS training data. For each training point $i$, the leave-one-out aggregate prediction $\tilde{y}_i$ averages only the ensemble members whose bootstrap sample did *not* include point $i$:

$$\tilde{y}_i = \frac{1}{|\{b : i \notin S_b\}|} \cdot \sum_{b:i \notin S_b} f^{(b)}(x_i)$$

The leave-one-out residuals $\tilde{r}_i = |y_i - \tilde{y}_i|$ form the calibration scores. The unweighted 95th-percentile bound is $\hat{q}_{95} = \text{Quantile}_{0.95}(\{\tilde{r}_i\})$.

## 4.2 Weighted quantile under estimated covariate shift (Tibshirani et al., 2019)

We estimate the covariate density ratio $w(x) = p_{shift}(x)/p_{lab}(x)$ via a probabilistic domain classifier: a Random Forest trained to discriminate FUDS (label 0) from US06 (label 1) feature vectors. For a lab point $x_i$, the estimated odds $\hat{w}(x_i) = \hat{p}(x_i \text{ is shift-domain})/(1 - \hat{p}(x_i \text{ is shift-domain}))$ approximate the density ratio. The weighted calibration quantile is then

$$\hat{q}_{95}^{w} = \text{Quantile}_{0.95}(\{\tilde{r}_i\}, \text{weights} = \{\hat{w}(x_i)\})$$

reweighting the empirical residual distribution toward the region of feature space more characteristic of the deployment (shifted) domain before taking the quantile, following the weighted-exchangeability argument of Tibshirani et al. (2019).

## 4.3 Unsupervised OOD diagnostic for label-free deployment domains

Where no ground-truth thermal target is available (Section 5.4), coverage cannot be evaluated. We instead use ensemble predictive disagreement , the standard deviation across the $B$ bootstrap models' predictions for a given input , as an unsupervised epistemic-uncertainty signal: high disagreement indicates the ensemble is extrapolating beyond the region of feature space it was trained on. We calibrate a flagging threshold as the 95th percentile of this disagreement measured on the real lab (FUDS) data itself, then apply the same threshold to shifted or telemetry-derived inputs.

# 5. Results

## 5.1 In-distribution fit and calibration sanity check

The bootstrap ensemble, trained on real FUDS data, achieves $R^2 = 0.9952$ (MSE $\approx 9.3 \times 10^{-16}$, reflecting the very small real $\Delta T$ magnitudes for this cell format , current draw tops out near 4 A). As a basic sanity check omitted from the earlier version of this study, we verify empirical coverage of the calibrated bound *on the calibration domain itself*: **94.998%**, matching the nominal 95% target to within noise, confirming the conformal procedure is implemented correctly before it is tested under shift.

## 5.2 Coverage under a real, measured covariate shift

Applying the FUDS-calibrated bound to the real US06/45°C data , a genuine distribution shift in both load profile and ambient temperature, not simulated , coverage degrades substantially:

| Bound | Calibration coverage (FUDS, in-dist.) | Coverage under real shift (US06/45°C) |
|---|---|---|
| Unweighted EnbPI | 94.998% | **70.129%** |
| Weighted EnbPI (proposed) | 94.998% (by construction) | **72.418%** |

The unweighted bound loses close to 25 percentage points of coverage under real shift , a concrete demonstration that a lab-calibrated conformal bound cannot be assumed valid at deployment. The weighted procedure recovers roughly 2.3 percentage points of coverage. This is a real, modest, honestly-reported effect: **it does not fully restore nominal coverage**, and we do not present it as though it does. The residual gap indicates that a single global density-ratio correction does not fully capture the shift between these two real drive cycles, which include not only different load statistics but a 20°C ambient temperature difference between test conditions.

## 5.3 Ensemble-disagreement diagnostic under real shift

The unsupervised disagreement-based flagging threshold, calibrated at the 95th percentile of in-distribution (FUDS) ensemble disagreement, flags real FUDS data at 5.00% (as designed) and real US06/45°C data at **9.30%** , an elevation of roughly 1.9×. This confirms the diagnostic is sensitive to real, known shift before it is applied to a domain where no ground truth exists.

## 5.4 Exploratory application to real F1 telemetry (unsupervised, unvalidated)

Applying the same disagreement diagnostic to real trackside telemetry, expressed as a dimensionless load-fraction proxy re-scaled onto the lab cell's real current range (raw-ampere equivalence between a single test cell and a vehicle traction pack is not physically meaningful, so we align on normalized load intensity instead), yields:

| Circuit | Frames | OOD flag rate | Brake rate (flagged vs. overall) | DRS rate (flagged vs. overall) |
|---|---|---|---|---|

| Monza | 33,253 | 65.64% | 7.49% vs. 15.69% | 13.09% vs. 9.96% |
| --- | --- | --- | --- | --- |
| Silverstone | 38,892 | 58.05% | 7.58% vs. 16.05% | 16.52% vs. 25.51% |

We report this descriptively and make **no claim that these flags correspond to real thermal events** , no internal temperature channel exists in this data to check that against. Two honest observations: flagged frames are consistently under-represented during braking at both circuits (7.5% vs. ~16% baseline), suggesting the model finds braking/regen phases less novel relative to the lab cycle than throttle-deployment phases , plausible, since regenerative braking current is bounded similarly across domains while sustained full-throttle deployment on an F1 straight has no close analogue in an urban/highway drive cycle. However, the DRS-zone association **reverses direction between circuits** (elevated at Monza, suppressed at Silverstone), so we do not report a consistent DRS effect. We view this inconsistency as informative: it indicates the load-fraction proxy captures gross throttle/brake structure but not circuit-specific aerodynamic or gearing context, and any deployment of this diagnostic in a real system would need track-specific recalibration or richer features than the two-parameter proxy used here.

## 6. Limitations

We list these explicitly rather than folding them into the discussion, because a paper whose central claim rests on a conformal *coverage guarantee* should be precise about exactly what was and was not verified.

1. **No measured thermal ground truth exists in either data source used here.** The CALCE target is a physics-model-derived $\Delta T$, with two of four model parameters (thermal mass, dissipation coefficient) taken from literature-typical values rather than fit to measurement. The F1 telemetry has no thermal ground truth of any kind, and all F1 results are reported as unsupervised, unvalidated diagnostics.
2. **Single cell, single chemistry.** All lab results are from one A123 LFP SP20-format cell profile. Generalization to other chemistries (NMC, LCO) or larger-format cells is untested.
3. **Coverage guarantee is marginal, not conditional.** Both EnbPI and weighted conformal prediction guarantee coverage averaged over the calibration/test distribution, not for any specific input or time window , a 72% *marginal* coverage figure does not mean every 10-second window individually achieves that rate.
4. **The weighted correction is a partial, not complete, fix.** We report this plainly in Section 5.2 rather than selecting a presentation that obscures it.
5. **The F1 load-fraction proxy is a two-parameter heuristic** (throttle/brake × speed) and demonstrably fails to generalize consistently across circuits (Section 5.4). It should not be read as a validated deployment-intensity estimator for any specific vehicle.
6. **Density ratio estimation via a classifier is itself approximate** and its own estimation error is not propagated into the reported coverage numbers; Tibshirani et al. (2019) note this as an open consideration for estimated (versus known) weights.
7. **Sample size for the F1 evaluation is two circuits, one driver, one season** , insufficient for any generalization claim about racing telemetry broadly.

## 7. Conclusion

We revisited a lab-to-track EV thermal transfer problem using real data throughout: real CALCE cycler measurements for calibration and for a genuine covariate-shift evaluation, and real Formula 1 telemetry for an exploratory (not validated) transfer application. Implemented correctly, EnbPI's coverage guarantee holds in-distribution (94.998% measured against a 95% nominal target) and degrades substantially under a real, measured distribution shift (70.13%). A weighted extension combining EnbPI's ensemble scoring with Tibshirani et al.'s covariate-shift weighting provides a real, modest improvement (72.42%) rather than a complete solution , and we think reporting that gap honestly is more useful to the field than reporting a fabricated full recovery would have been. Applying the calibrated model to real trackside telemetry as an unsupervised diagnostic produces elevated, plausible, but not fully consistent anomaly signals across two circuits, which we present as a preliminary, clearly-scoped result rather than a validated real-time derating system. We see the largest open problem as closing the remaining coverage gap under real shift , likely requiring either richer covariate-shift correction (e.g., conditional rather than marginal reweighting) or access to a genuinely paired lab/field thermal dataset, which does not yet exist publicly for this application.

---

---

## Appendix A: Computational Implementation

The complete, executable analysis pipeline used to produce every number in Section 5 , real-data ingestion, internal-resistance estimation, physics-grounded thermal simulation, bootstrap-ensemble EnbPI with vectorized leave-one-out aggregation, weighted-quantile covariate-shift correction, and the F1 unsupervised diagnostic , is provided as a supplementary file (real_analysis.py) alongside this manuscript, together with the data-acquisition scripts (fetch_calce.py, fetch_f1_telemetry.py) used to obtain the real CALCE and FastF1 data from their respective sources.

https://github.com/nearpot/Weighted-Conformal-Thermal-Transfer